\documentclass[11pt]{article}

\usepackage[final]{acl}

\usepackage{times}
\usepackage{latexsym}
\usepackage{url}
\usepackage{enumitem}
\usepackage{amsmath}
\usepackage{booktabs}
\usepackage{array} 
\usepackage{float}
\usepackage{multirow}
\usepackage{algorithm}
\usepackage{algorithmic}
\usepackage{CJKutf8} 
\usepackage{fontspec}

\usepackage[T1]{fontenc}

\usepackage[utf8]{inputenc}

\usepackage{microtype}

\usepackage{inconsolata}

\usepackage{graphicx}

\title{Enhancing Cross-Lingual Transfer through Reversible Transliteration: A Huffman-Based Approach for Low-Resource Languages}

\author{
  \textbf{Wenhao Zhuang\textsuperscript{1,2}},
  \textbf{Yuan Sun\textsuperscript{1,2,3,*}},
  \textbf{Xiaobing Zhao\textsuperscript{1,2,3}}\\
  \textsuperscript{1}Minzu University of China, Beijing, China\\
  \textsuperscript{2}National Language Resource Monitoring \& Research Center Minority Languages Branch\\
  \textsuperscript{3}Institute of National Security, Minzu University of China, Beijing, China\\
  \\
  Emails: \texttt{sdrz\_zwh@163.com}, 
  \texttt{sunyuan@muc.edu.cn},
  \texttt{nmzxb\_cn@163.com}\\
  {* Corresponding author: Yuan Sun}
}

\begin{document}
\maketitle
\begin{abstract}


As large language models (LLMs) are trained on increasingly diverse and extensive multilingual corpora, they demonstrate cross-lingual transfer capabilities. However, these capabilities often fail to effectively extend to low-resource languages, particularly those utilizing non-Latin scripts. While transliterating low-resource languages into Latin script presents a natural solution, there currently lacks a comprehensive framework for integrating transliteration into LLMs training and deployment. Taking a pragmatic approach, this paper innovatively combines character transliteration with Huffman coding to design a complete transliteration framework. Our proposed framework offers the following advantages: 1) Compression: Reduces storage requirements for low-resource language content, achieving up to 50\% reduction in file size and 50-80\% reduction in token count. 2) Accuracy: Guarantees 100\% lossless conversion from transliterated text back to the source language. 3) Efficiency: Eliminates the need for vocabulary expansion for low-resource languages, improving training and inference efficiency. 4) Scalability: The framework can be extended to other low-resource languages. We validate the effectiveness of our framework across multiple downstream tasks, including text classification, machine reading comprehension, and machine translation. Experimental results demonstrate that our method significantly enhances the model's capability to process low-resource languages while maintaining performance on high-resource languages. Our data and code are publicly available at \url{https://github.com/CMLI-NLP/HuffmanTranslit}.

\end{abstract}

\section{Introduction}

Large language models have demonstrated remarkable multilingual transfer capabilities, enabling knowledge transfer from one language to another without additional training~\cite{qi2023cross, gao2024multilingual, ye2023language}. However, this transfer ability often performs poorly in low-resource languages, primarily constrained by three factors: scarcity of training data~\cite{costa2022no}, insufficient cross-lingual word embedding alignment~\cite{deshpande2021bert}, and writing system differences~\cite{anastasopoulos2019pushing, muller2021being}.

Common approaches to improving LLMs' adaptability to low-resource languages include continued pre-training and supervised fine-tuning~\cite{tao2024unlocking}. Due to the low representation of low-resource languages in tokenizers and frequent occurrence of UNKnown tokens~\cite{moosa2023does}, vocabulary expansion becomes a primary task~\cite{zhuang2025cute}. However, ensuring high performance for multiple low-resource languages is extremely challenging, facing two key issues. The first issue is the increased training and inference costs due to vocabulary size, as vocabulary must inevitably expand with the addition of languages to ensure tokenization performance for each language~\cite{purkayastha2023romanization}. The second issue is the curse of multilinguality, which means that using a fixed-capacity model to pre-train multiple languages can improve cross-lingual performance to some extent, but performance begins to decline beyond a certain point~\cite{conneau2019unsupervised}.

To improve cross-lingual transfer while avoiding issues associated with extensive vocabulary expansion, transliteration of low-resource languages has emerged as a viable approach. Transliteration refers to the process of converting text from one writing system to another according to specific rules, typically converting non-Latin scripts to Latin alphabet representation ~\cite{wellisch1978conversion}. Previous studies have demonstrated that transliterating text into a common character set can enhance cross-lingual transfer performance for low-resource languages with non-Latin scripts ~\cite{liu2024translico, liu2025transliterations}. This improvement is attributed to the common character set facilitating knowledge transfer through lexical overlap ~\cite{dhamecha2021role, pires2019multilingual, amrhein2020romanization} and enabling the reuse of existing information in embedding matrices ~\cite{purkayastha2023romanization}. Appending transliterated content to prompt templates for low-resource languages has been shown to improve downstream task performance ~\cite{ma2024exploring}.

Most research on transliteration relies on existing tools like UROMAN~\cite{hermjakob2018out}, which maps any UTF-8 character to Latin letters, to investigate the impact of transliteration on cross-lingual transfer or alignment. However, this transliteration process is irreversible; due to the potential loss of characteristic information from the original script during transliteration, it is impossible to accurately restore the transliterated Latin letters back to the source language script, which limits its practical applications~\cite{amrhein2020romanization}. Furthermore, low-resource languages typically utilize extended Unicode character sets for encoding, resulting in their textual data occupying more storage space compared to languages like English. This storage overhead issue becomes more prominent when processing large-scale multilingual corpora. Therefore, this paper focuses on two key issues: how to implement a reversible transliteration mechanism to facilitate practical applications while maintaining cross-lingual transfer effectiveness, and how to achieve text compression during the transliteration process to facilitate storage and training. We observe that these two points correspond precisely to the reversibility and compression properties of Huffman coding, which provides the theoretical foundation for our Huffman coding-based transliteration scheme.

We selected three low-resource languages: Tibetan, Mongolian, and Uyghur, which are minority languages in China with a total user base exceeding 30 million speakers, along with English and Chinese as high-resource languages for our experiments. We conducted continued pre-training of open-source LLMs using corpora obtained through various transliteration methods, analyzed cross-lingual transfer performance across text classification, named entity recognition, machine reading comprehension, knowledge extraction, and machine translation tasks, while also comparing compression rates among different transliteration methods. To enable the model to directly serve low-resource language users, we developed a FastText-based automatic transliteration framework that performs language detection before and after model processing, implementing transliteration and restoration of input and output, thereby maintaining native language interaction at the user end. In summary, our contributions are as follows:

\begin{itemize}[topsep=0pt,itemsep=0pt]
  \item We propose a Huffman coding-based transliteration scheme for low-resource languages, achieving reversibility in the transliteration process and addressing the limitations of traditional transliteration methods in practical applications.
  \item Leveraging the compression properties of Huffman coding, we effectively reduce the storage overhead of low-resource language texts, making the training of large-scale multilingual corpora more efficient.
  \item We develop an end-to-end framework integrating FastText language identification, enabling automatic transliteration and restoration of low-resource languages while maintaining native language interaction and improving performance across multiple downstream tasks.
\end{itemize}

\section{Related Works}




\paragraph{Cross-lingual Transfer for Low-resource Languages}
To enhance low-resource language performance in LLMs, one primary approach uses continued pre-training and supervised fine-tuning with low-resource corpora~\cite{wenhao2024tilamb}. However, this method faces significant challenges: it requires complex vocabulary expansion and model architecture modifications, resulting in poor scalability, and most critically, suffers from limited training data availability~\cite{joshi2020state}. 

Alternative approaches focus on improving cross-lingual transfer capabilities through various mechanisms: concatenating multilingual input sequences to leverage shared representation spaces~\cite{kim2024translating,tanwar2023multilingual,cueva2024adaptive}, projecting target language representations onto high-resource languages for enhanced feature extraction~\cite{xu2023language}, and increasing the parallel content in multilingual training corpora~\cite{zhuang2025cute}.

While these methods aim to transfer capabilities from resource-rich to low-resource languages, a fundamental challenge remains: the substantial differences in writing systems among low-resource languages. Unifying multiple languages into a single writing system could potentially address vocabulary challenges and promote vocabulary sharing, thereby facilitating cross-lingual knowledge transfer~\cite{purkayastha2023romanization}.




\paragraph{Tokenization and Vocabulary Expansion}
Existing subword tokenizers (such as BPE~\cite{sennrich2015neural} and SentencePiece~\cite{kudo2018sentencepiece}) have been widely adopted for low-resource languages. However, due to the limited representation of these languages in pre-training corpora, they suffer from insufficient vocabulary coverage, over-segmentation, and high ratios of unknown tokens. While vocabulary expansion~\cite{cui2023efficient} offers a potential solution, it introduces new challenges: the need for substantial training data to adequately train new tokens, and increased model capacity requirements to mitigate the multilingual curse~\cite{conneau2019unsupervised}.

Recent approaches have focused on more efficient solutions, such as leveraging shared linguistic information and cross-lingual word embedding alignment~\cite{ogueji2021small,liu2021mulda}, which improve tokenization without significant vocabulary expansion. Notably, transliterating low-resource languages into a unified writing system has shown promising results~\cite{dhamecha2021role,liu2024transmi}, simultaneously enhancing vocabulary sharing and model transfer capabilities while avoiding the computational overhead of vocabulary expansion.




\paragraph{Romanization and Transliteration}
Romanization is the process of mapping various characters to Latin characters, though this process is typically irreversible. Its objective is to approximate the pronunciation of the original character text as closely as possible. Specialized tools like UROMAN~\cite{hermjakob2018out} can romanize almost all characters by directly mapping UTF-8 characters to Latin letters, though this process involves information loss, such as the omission of tonal information. There are also general character conversion tools like uconv that can preserve more original character information, such as adding diacritical marks, but this limits subword sharing across languages.

The Tibetan, Mongolian, Uyghur, and Chinese languages used in our experiments can all be romanized through UROMAN; however, due to the uniqueness of the Tibetan writing system, uconv currently cannot transliterate Tibetan. Romanization encoding has been studied in both natural language processing and speech processing domains, such as its application in multilingual pre-trained language models to enhance low-resource languages~\cite{purkayastha2023romanization}, and in speech processing systems' pre-training as additional forced alignment for text labeling~\cite{pratap2024scaling}. Moreover, phonological distinctions may be lost during romanization - for instance, Chinese characters become toneless pinyin when romanized, with a single pinyin potentially corresponding to many different characters. UROMAN also converts numbers from different writing systems into Western Arabic numerals~\cite{ding2024romanization}, which further complicates the process of converting romanized text back to source languages, particularly when users expect LLMs to output in their native writing systems. In contrast, our proposed Huffman coding-based transliteration method is an innovative approach that balances transliteration (improving cross-lingual transfer), compression (reducing storage and training costs), and reversibility (facilitating practical restoration and interaction).

\section{Methodology}

\subsection{Overview}

\begin{figure*}[ht]
\centering
\includegraphics[width=\textwidth]{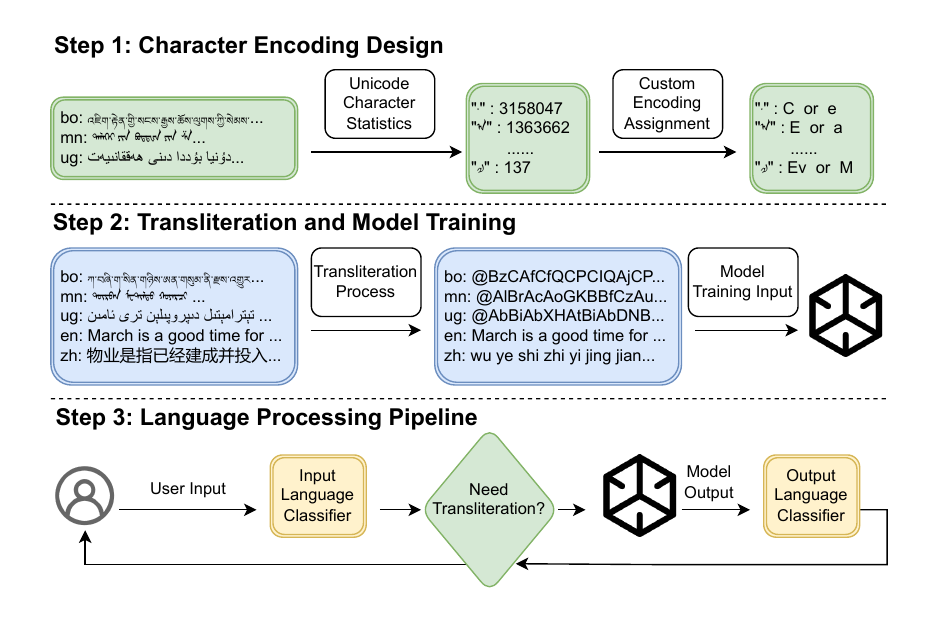}
\caption{Overview of our three-stage approach. Step 1: Character encoding design with Unicode character statistics and custom encoding assignment. Step 2: Transliteration process for model training input. Step 3: Language processing pipeline with language classification for user interaction.}
\label{fig:framework}
\end{figure*}

We propose a three-stage processing approach for low-resource language transliteration and applications: (1) character encoding design: analyzing character frequencies and designing custom encodings. (2) transliteration and model training: training on transliterated raw corpora. (3) end-to-end language processing pipeline: comprising input language classification and processing, model inference, and output language classification and processing, as shown in Figure~\ref{fig:framework}.

In this study, we focus on three low-resource languages of China: Tibetan, Uyghur, and Mongolian (see Table~\ref{tab:language_info}). These languages represent different writing systems, with a combined user base exceeding 30 million speakers. We select these languages because they present significant challenges for cross-lingual transfer: they use non-Latin scripts, have limited digital resources, and exhibit distinct writing systems that differ substantially from high-resource languages like Chinese and English.

\begin{table}[h]
\small
\centering
\begin{tabular}{c|c|c}
\toprule
\textbf{Name} & \textbf{ISO 639-1} & \textbf{Writing System}  \\
\midrule
Tibetan & \texttt{bo}  & Tibetan script \\
Uyghur & \texttt{ug} & Uyghur Arabic script \\
Mongolian & \texttt{mn} & Traditional Mongolian script \\
\bottomrule
\end{tabular}
\caption{Overview of the low-resource languages studied in this work.}
\label{tab:language_info}
\end{table}

\subsection{Character Frequency Analysis}


Since English already uses Latin script and Chinese has well-established romanization tools for converting characters to pinyin, we focused our transliteration efforts on three low-resource languages: Uyghur, Tibetan, and Mongolian. Our character frequency analysis began with the CUTE open-source parallel dataset ~\cite{zhuang2025cute}, which provided aligned text across Chinese, Uyghur, Tibetan, and English. To incorporate Mongolian, which wasn't originally included in CUTE, we followed the dataset's methodology to collect and evaluate Mongolian translations, using human evaluators to assess translation quality from Chinese to Mongolian. For comprehensive character analysis, we sampled 3,000 instances from each of the three low-resource languages and conducted a thorough examination of their Unicode characters and frequencies. Through a systematic approach combining Unicode code point ranges, character naming conventions, and expert linguistic validation, we identified the core character sets for each language: 45 characters for Mongolian, 38 for Uyghur, and 81 for Tibetan. These character sets were then consolidated and arranged in descending order of frequency, providing a foundation for our transliteration scheme.

\subsection{Huffman-based Encoding Design}
\paragraph{Properties of Huffman Coding}

Huffman coding, as a variable-length encoding method, possesses both variable-length allocation and prefix properties. The variable-length allocation ensures that high-frequency characters receive shorter codes while low-frequency characters receive longer codes, providing a theoretical foundation for our text compression. The prefix property guarantees that no code is a prefix of any other character's code, facilitating unambiguous decoding.
\paragraph{Customized Encoding Scheme}

Based on the principles of Huffman coding, we designed an improved encoding scheme. To accommodate more languages, our scheme, while not strictly adhering to the prefix property, ensures unambiguous decoding through structured design. Specifically, we constrain all codes to follow the pattern "First letter capitalized, subsequent letters lowercase" and employ a maximum matching strategy for decoding. Table \ref{tab:encoding_pattern} demonstrates the possibilities of this design pattern at different lengths.

\begin{table}[H]

\centering
\begin{tabular}{llr}
\toprule
\textbf{Length} & \textbf{Pattern} & \textbf{Capacity} \\
\midrule
One & A, B, ..., Z & 26 \\
Two & Aa, Ab, ..., Zz & 676 \\
Three & Aaa, Aab, ..., Zzz & 17,576 \\
Four & Aaaa, Aaab, ..., Zzzz & 456,976 \\
\midrule
\multicolumn{2}{l}{Total (up to four characters)} & 475,254 \\
\bottomrule
\end{tabular}
\caption{Encoding patterns and theoretical capacity for different lengths. The pattern consists of one uppercase letter followed by zero or more lowercase letters.}
\label{tab:encoding_pattern}
\end{table}

This design offers three key advantages: (1) By constraining the first character to be uppercase and subsequent characters to be lowercase, combined with the maximum matching strategy, it ensures unambiguous decoding. (2) It maintains the core principle of Huffman coding, allowing for variable-length code allocation based on character frequencies. (3) It provides significant scalability, theoretically supporting encoding for up to 475,254 characters. In this study, we implemented a subset of the two-character scheme, utilizing 21 single-character codes (B-X) and 141 two-character codes (Aa-Fk), totaling 162 encoding options. This scale is sufficient to cover the character sets of Uyghur, Tibetan, and Mongolian languages.

\subsection{Transliteration Strategies}
\label{sec:trans_strategies}






To explore optimal transliteration strategies, we designed three progressive transliteration schemes, with each scheme building upon and improving its predecessor.
\paragraph{Basic Transliteration Strategy}
We designed a transliteration scheme using "First-letter-capitalized + lowercase" encoding rules. This scheme transliterates Uyghur, Tibetan, and Mongolian into Latin alphabet representations according to encoding rules, while converting Chinese characters into Pinyin and preserving English text unchanged. While this strategy achieved basic transliteration functionality and reversibility, it did not account for the tokenizer's characteristics, leaving room for further token compression.
\paragraph{Tokenizer-based Optimization Strategy}
To address the token optimization potential in the basic strategy, we analyzed the characteristics of the Llama2 tokenizer. Research showed that 66\% of original characters required four tokens for representation. In response, we innovatively utilized single tokens from the Llama2 tokenizer~\cite{touvron2023llama} as encoding mappings for original characters, enabling all 162 original characters to be represented by single tokens. The comparative token distribution is shown in Table \ref{tab:token_distribution}.

\begin{table}[H]

\centering
\setlength{\tabcolsep}{2pt}
\begin{tabular}{lrrrr}
\toprule
\textbf{Method} & \textbf{1-token} & \textbf{2-token} & \textbf{3-token} & \textbf{4-token} \\
\midrule
Original & 1 & 45 & 9 & 107 \\
Basic & 90 & 72 & 0 & 0 \\
Optimized & 162 & 0 & 0 & 0 \\
\bottomrule
\end{tabular}
\caption{Character distribution by token length after Llama2 tokenization. The Optimized method achieves single-token encoding for all characters.}
\label{tab:token_distribution}
\end{table}

\paragraph{Hybrid Vocabulary Strategy}
Building upon the second strategy, we leveraged the linguistic patterns inherent in the transliterated text to train a specialized vocabulary of 4,000 tokens. This vocabulary was merged with Llama2's original 32,000-token vocabulary to create a hybrid vocabulary of 33,738 tokens, maintaining efficient single-character encoding while capturing common character combinations. The comparison of the three strategies is presented in Table \ref{tab:vocab_comparison}. For a comprehensive analysis of file size and token compression ratios across different languages and strategies, see Appendix \ref{sec:compression_analysis}.

\begin{table}[H]
\centering
\setlength{\tabcolsep}{6pt}
\begin{tabular}{lrrr}
\toprule
\textbf{Strategy} & \textbf{Vocab Size} & \textbf{Compr.} & \textbf{Cost} \\
\midrule
Basic & 32,000 & 1.63× & Low \\
Tokenizer & 32,000 & 2.35× & Medium \\
Hybrid & 33,738 & 3.04× & High \\
\bottomrule
\end{tabular}
\caption{Comparison of different strategies. Compr. shows average token compression ratio across Tibetan, Mongolian and Uyghur languages.}
\label{tab:vocab_comparison}
\end{table}

\subsection{Reversibility Mechanism}



Our transliteration system achieves perfect reversibility through a carefully designed mapping mechanism, maintaining bidirectional mappings between original characters and Latin codes. For characters not in the mapping tables (e.g., emojis, rare characters, or special symbols), the system preserves them using '@' markers with proper escape sequences (e.g., '@@' for the '@' character itself), ensuring no information loss during transliteration. The detailed process is shown in Algorithm~\ref{alg:transliteration}.

\begin{algorithm}[t]
\caption{Reversible Transliteration System}
\label{alg:transliteration}
\begin{algorithmic}
\STATE \textbf{Input:} $M_{c2l}, M_{l2c}$: Bidirectional mappings, $text$: Input text
\STATE \textbf{Output:} Transliterated or restored text
\STATE
\STATE \textbf{Function} ToLatin($text$):
\STATE \hspace{0.5cm} $result \leftarrow []$
\STATE \hspace{0.5cm} \textbf{for} each $c$ in $text$ \textbf{do}
\STATE \hspace{1.0cm} Append $M_{c2l}[c]$ if exists, else preserve as @...@
\STATE \hspace{0.5cm} \textbf{end for}
\STATE \hspace{0.5cm} \textbf{return} joined result
\STATE
\STATE \textbf{Function} FromLatin($latin\_text$):
\STATE \hspace{0.5cm} $result \leftarrow []$, $i \leftarrow 0$
\STATE \hspace{0.5cm} \textbf{while} $i <$ length($latin\_text$) \textbf{do}
\STATE \hspace{1.0cm} Process @ markers or find longest matching code
\STATE \hspace{1.0cm} Advance $i$ accordingly
\STATE \hspace{0.5cm} \textbf{end while}
\STATE \hspace{0.5cm} \textbf{return} joined result
\end{algorithmic}
\end{algorithm}

The system employs a greedy matching strategy during restoration, where it attempts to match the longest possible Latin code sequence for mapped characters while correctly handling preserved sequences between '@' markers. This dual mechanism ensures 100\% restoration accuracy by either mapping characters through the bidirectional tables or preserving them in their original form. 

\subsection{Auxiliary Models for Practical Deployment}

To achieve end-to-end system deployment, we developed three auxiliary models. At the input stage, we trained a FastText-based classifier specifically for identifying Mongolian, Tibetan, Uyghur, Chinese, and other languages. At the output stage, we trained a FastText language classifier tailored to the characteristics of transliterated text to guide language restoration. Additionally, to accurately handle the conversion from Chinese pinyin to characters, we fine-tuned a specialized model based on Qwen2.5-0.5B~\cite{yang2024qwen2}. The detailed training processes and evaluation results of these models are presented in Appendix~\ref{sec:Auxiliary Models}.

\section{Experiments and Analysis}
\begin{table*}[ht]
\centering
\resizebox{\textwidth}{!}{%
\setlength\tabcolsep{6pt}
\begin{tabular}{l|ccc|cc|cc}
\toprule
\multirow{2}{*}{\textbf{Model}} & \multicolumn{3}{c|}{\textbf{Low-resource Languages (Acc / F1)}} & \multicolumn{2}{c|}{\textbf{Chinese}} & \multicolumn{2}{c}{\textbf{Average}} \\
& \texttt{bo} & \texttt{mn} & \texttt{ug} & Acc & F1 & Minorities & All \\
\midrule
Base Llama2 & 28.65 / 21.23 & 1.78 / 1.65 & 73.33 / 74.69 & 86.12 & 85.91 & 13.48 / 9.01 & 48.15 / 48.78 \\
Expanded Vocab & \underline{53.96} / 51.69 & 64.45 / \underline{67.95} & 76.00 / 82.91 & 89.95 & 89.91 & \underline{62.58} / \underline{64.56} & \underline{75.64} / \underline{76.34} \\
UROMAN & 49.37 / 48.88 & \underline{64.92} / 67.78 & \textbf{81.33} / \textbf{86.18} & 89.82 & 89.75 & 62.10 / 63.47 & 75.33 / 75.84 \\
Basic Trans. & 52.16 / 52.18 & \textbf{66.46} / \textbf{69.55} & 74.67 / 81.51 & 89.92 & 89.81 & \textbf{63.40} / \textbf{65.17} & \textbf{76.06} / \textbf{76.60} \\
Token-Opt Trans. & \textbf{54.14} / \textbf{54.94} & 61.25 / 64.61 & \underline{81.00} / \underline{85.31} & \textbf{90.15} & \textbf{90.07} & 60.80 / 63.54 & 74.81 / 75.86 \\
Hybrid Trans. & 50.45 / \underline{52.58} & 61.25 / 65.15 & 65.33 / 75.59 & \underline{90.00} & \underline{89.95} & 58.80 / 62.15 & 73.68 / 75.02 \\
\bottomrule
\end{tabular}%
}
\caption{Performance comparison on the WCM-v2 dataset. The best scores are in \textbf{bold}, with the second best \underline{underlined}. Base Llama2: directly fine-tuned on original texts; Expanded Vocab: vocabulary expansion for each low-resource language; Basic/Token-Opt/Hybrid Trans.: three progressive transliteration strategies. Minorities average is calculated as the mean of scores for low-resource languages.}
\label{tab:classification_results}
\end{table*}

\begin{table*}[ht]
\centering
\small
\setlength\tabcolsep{12pt}  
\begin{tabular}{l|cc|cc|cc|cc}
\toprule
\multirow{2}{*}{\textbf{Model}} & \multicolumn{4}{c|}{\textbf{CMRC-Trained}} & \multicolumn{4}{c}{\textbf{SQuAD-Trained}} \\
\cmidrule(lr){2-5} \cmidrule(l){6-9}
& \multicolumn{2}{c|}{Chinese} & \multicolumn{2}{c|}{Tibetan} & \multicolumn{2}{c|}{English} & \multicolumn{2}{c}{Tibetan} \\
& EM & F1 & EM & F1 & EM & F1 & EM & F1 \\
\midrule
Base Llama2 & 77.2 & 89.5 & 7.9 & 45.8 & \underline{89.5} & \underline{95.3} & 6.5 & 50.8 \\
Expanded Vocab & 81.3 & 91.1 & 11.5 & 50.6 & \textbf{89.9} & \textbf{95.7} & 10.3 & 58.9 \\
UROMAN & 79.6 & 88.3 & 12.0 & 53.4 & 84.2 & 88.7 & 11.0 & 61.1 \\
Basic Trans. & \underline{87.7} & \underline{92.7} & \underline{15.5} & \underline{59.5} & 87.7 & 89.1 & \underline{12.7} & \underline{65.1} \\
Token-Opt Trans. & \textbf{88.4} & \textbf{93.6} & \textbf{16.0} & \textbf{60.2} & 88.0 & 89.3 & \textbf{13.5} & \textbf{66.5} \\
Hybrid Trans. & 83.1 & 90.2 & 14.8 & 58.8 & 87.2 & 89.0 & 12.3 & 64.9 \\
\bottomrule
\end{tabular}
\caption{Machine Reading Comprehension performance comparison. The best scores are in \textbf{bold}, with the second best \underline{underlined}. Results show both source language (Chinese/English) and target language (Tibetan) performance under different training settings. EM: Exact Match score; F1: F1 score.}
\label{tab:tibetan_qa}
\end{table*}


To evaluate the effectiveness of different transliteration schemes, we conducted a series of experiments examining the cross-lingual transfer performance of models trained with various transliteration strategies. We specifically focused on whether the models could effectively transfer knowledge to low-resource languages (Tibetan, Mongolian, and Uyghur) while maintaining performance in high-resource languages (Chinese and English).

\subsection{Experimental Setup}




We adopt the following experimental procedure: First, we process the pre-training corpus using different transliteration methods, followed by continued pre-training of the model. The choice of continued pre-training over training from scratch is motivated by the common challenge of insufficient training data faced by low-resource languages, which makes it difficult to support a complete pre-training process. After pre-training, we perform supervised fine-tuning using downstream task data from high-resource languages, and then directly conduct zero-shot evaluation on low-resource languages to verify the model's cross-lingual transfer capability. 

For Tibetan, Uyghur, and Mongolian languages, we identified a limited number of available datasets. Our experiments encompassed three primary tasks: text classification, machine reading comprehension, and translation. These tasks evaluated the model's capabilities across different levels of language processing, thereby enabling a comprehensive assessment of the transliteration scheme's effectiveness. Details regarding the pre-training data and parameter settings can be found in Appendix~\ref{sec:training_details}.

We designed the following comparative experiments:
\begin{itemize}
    \item Direct Continued Pre-training: Continuing pre-training on the original model using raw corpora.
    \item Vocabulary Expansion: Augmenting the original model's vocabulary with dedicated lexicons for each low-resource language.
    \item UROMAN Transliteration: Applying universal romanization tools for transliteration.
    \item Three Progressive Transliteration Strategies: As detailed in Section~\ref{sec:trans_strategies}.
\end{itemize}

\subsection{Text Classification}


We first evaluated the effectiveness of various transliteration schemes on the text classification task. The experiments utilized the WCM-v2 dataset (see Appendix~\ref{sec:WCM_v2_details}), a classification dataset encompassing multiple ethnic minority languages of China~\cite{yang2022cino}. This dataset maintains balanced distributions across both categories and languages, containing texts from 10 domains including arts, geography, and history. The experimental results are shown in Table~\ref{tab:classification_results}. To comprehensively evaluate the effectiveness of each approach, we focus not only on the overall performance but also specifically on the average performance across low-resource languages.

\subsection{Machine Reading Comprehension}

For the machine reading comprehension task, we evaluate the models' performance by fine-tuning them on the Chinese CMRC dataset~\cite{cui2019span} and English SQuAD dataset~\cite{rajpurkar2016squad}, followed by zero-shot testing on the TibetanQA dataset~\cite{sun2021construction} to assess their cross-lingual transfer capabilities. We also report the performance on the source languages to verify that our approaches maintain strong performance on high-resource languages while enabling effective cross-lingual transfer. The results are shown in Table~\ref{tab:tibetan_qa}. For detailed information about the datasets, please refer to Appendix~\ref{sec:MRC_details}.

\subsection{Machine Translation}


To evaluate the models' machine translation capabilities for low-resource languages, we conduct experiments on Chinese-to-Tibetan (zh-bo) and Chinese-to-Uyghur (zh-ug) translation tasks using the Flores-200 dataset~\cite{costa2022no}. We employ few-shot prompting with three carefully selected examples for each language pair, ensuring the examples cover diverse linguistic patterns. The evaluation uses three standard metrics: BLEU score for overall translation quality, chrF for character-level accuracy, and Translation Edit Rate (TER) for measuring the amount of editing required to match the reference translation. Table~\ref{tab:translation_results} presents the results of our comparative evaluation. For detailed information about the dataset and prompts used, please refer to Appendix~\ref{sec:translation_details}.

\begin{table*}[ht]
\centering
\setlength\tabcolsep{12pt}
\begin{tabular}{l|ccc|ccc}
\toprule
\multirow{2}{*}{\textbf{Model}} & \multicolumn{3}{c|}{\textbf{Chinese-to-Tibetan (zh-bo)}} & \multicolumn{3}{c}{\textbf{Chinese-to-Uyghur (zh-ug)}} \\
& \textbf{BLEU}$\uparrow$ & \textbf{chrF}$\uparrow$ & \textbf{TER}$\downarrow$ & \textbf{BLEU}$\uparrow$ & \textbf{chrF}$\uparrow$ & \textbf{TER}$\downarrow$ \\
\midrule
Base Llama2      & 3.5 & 0.28 & 0.92 & 4.2 & 0.31 & 0.89 \\
Expanded Vocab   & 5.0 & 0.35 & 0.86 & 5.7 & 0.38 & 0.83 \\
UROMAN           & 4.5 & 0.33 & 0.88 & 5.2 & 0.36 & 0.85 \\
Basic Trans.     & \underline{5.7} & \underline{0.37} & \underline{0.84} & \underline{6.4} & \underline{0.40} & \underline{0.81} \\
Token-Opt Trans. & \textbf{6.3} & \textbf{0.39} & \textbf{0.82} & \textbf{7.0} & \textbf{0.42} & \textbf{0.79} \\
Hybrid Trans.    & 3.8 & 0.30 & 0.90 & 4.5 & 0.33 & 0.87 \\
\bottomrule
\end{tabular}
\caption{Machine Translation performance comparison on Flores-200 dataset using few-shot prompting (3 examples). $\uparrow$: higher is better, $\downarrow$: lower is better. The best scores are in \textbf{bold}, with the second best \underline{underlined}. TER: Translation Edit Rate.}
\label{tab:translation_results}
\end{table*}

\subsection{Overall Analysis}
\paragraph{Cross-task Performance Analysis}
Through a comparative analysis of experimental results across text classification, machine reading comprehension, and machine translation tasks, our proposed transliteration approach demonstrated excellent cross-lingual transfer capabilities. In text classification tasks, the basic transliteration strategy achieved an average accuracy of 63.40\% on low-resource languages, showing an improvement of 0.82\% compared to the vocabulary expansion approach. For machine reading comprehension tasks, the tokenizer-optimized transliteration strategy achieved an exact match score of 16.0\% on Chinese-to-Tibetan transfer, outperforming the vocabulary expansion approach by 4.5\%. This strategy also exhibited superior performance in translation tasks, achieving a BLEU score of 6.3 in Chinese-to-Tibetan translation. Notably, these improvements were achieved while maintaining high performance on resource-rich languages, as exemplified by our approach achieving 90.15\% accuracy on Chinese text classification tasks.

\paragraph{Key Findings} Our experimental results yield three significant findings:
\begin{itemize}
    \item \textbf{Performance and Efficiency:} Our transliteration approaches consistently outperformed traditional vocabulary expansion methods across tasks, with both basic and tokenizer-optimized strategies showing exceptional results. By leveraging existing tokenizer characteristics, these approaches significantly improved low-resource language processing without vocabulary expansion.
    \item \textbf{Untapped Potential:} Despite using simple frequency-based encoding schemes (B-X, Aa-Fk) and random token assignments, our methods demonstrated remarkable effectiveness. This suggests substantial room for improvement through the incorporation of linguistic features and more sophisticated encoding strategies.
    \item \textbf{Scalable Framework:} Our findings establish a new paradigm for low-resource language processing, offering a more promising direction than vocabulary expansion. The success of this relatively simple implementation particularly demonstrates its potential for scaling to multiple low-resource languages.
\end{itemize}
These results not only validate our approach but also indicate that more sophisticated versions of these strategies could yield even more significant improvements in low-resource language processing.

\section{Conclusion}


In this paper, we introduce a novel Huffman-based transliteration framework that addresses three critical challenges in low-resource language processing: cross-lingual transfer, storage efficiency, and practical deployment. Our framework demonstrates superior performance across diverse tasks while maintaining a lightweight implementation. The basic and tokenizer-optimized strategies consistently outperform traditional approaches, achieving up to 4.5\% improvement in cross-lingual machine reading comprehension and significant gains in translation tasks, all while preserving performance on high-resource languages. Beyond performance gains, our approach offers unique advantages in compression efficiency, reducing both file size and token count by 2-3 times without sacrificing reversibility. Most importantly, our framework's success with simple frequency-based encoding suggests substantial potential for improvement through the incorporation of linguistic features and more sophisticated encoding strategies. These findings establish a promising direction for scaling language technologies to the world's many low-resource languages, offering a more practical alternative to the traditional vocabulary expansion paradigm.

\section*{Limitations}


While our approach demonstrates promising results, there are several important limitations to consider. Our current evaluation scope is restricted to three low-resource languages with non-Latin scripts. Although the framework is theoretically extensible to other writing systems, specific adaptations may be necessary to accommodate their unique characteristics. The limited availability of evaluation datasets for low-resource languages also poses a challenge, particularly in tasks like machine reading comprehension, where we could only assess performance on a subset of languages.

From a practical perspective, our approach faces a trade-off between storage efficiency and computational overhead. While we achieve significant reductions in storage requirements, the transliteration and restoration processes introduce additional computational steps that could impact real-time performance, especially in scenarios requiring frequent language switching. Furthermore, our current encoding scheme relies primarily on character frequency, leaving room for potential improvements through the incorporation of linguistic features such as phonemes and morphological information.

\section*{Acknowledgments}
This work is supported by the National Social Science Foundation (22\&ZD035), the National Nature Science Foundation (61972436), and
the Minzu University of China Foundation (GRSCP202316, 2023QNYL22, 2024GJYY43).

\bibliography{custom}

\clearpage
\appendix

\section{Auxiliary Models}
\label{sec:Auxiliary Models}


To achieve a comprehensive end-to-end system, we developed three auxiliary models for input language identification, output language identification, and Chinese pinyin conversion. These models collectively form a complete language processing pipeline, ensuring that the system can accurately process multilingual inputs and generate appropriate outputs.

\subsection{Input Language Classifier}

We trained specialized language identification models based on the FastText framework to accurately identify the language type of input text. The classifier supports five language categories: Tibetan (bo), Mongolian (mn), Uyghur (ug), Chinese (zh), and other languages (other). The training data was sourced from multiple datasets, including CUTE~\cite{zhuang2025cute}, WCM-v2~\cite{yang2022cino}, and other open-source datasets, to ensure the model can process text from various domains. The training parameters for the input language classifier are shown in Table~\ref{tab:classifier_params}.


The evaluation results on the test set of 5,000 entries show that the classifier achieved a high level of classification across all languages and can be used for actual classification needs. The evaluation results are shown in Table~\ref{tab:input_classifier_performance}.

\begin{table*}[t]
\centering
\setlength\tabcolsep{8pt}
\begin{tabular}{lrrl}
\toprule
\textbf{Parameter} & \textbf{Input Classifier} & \textbf{Output Classifier} & \textbf{Note} \\
\midrule
Learning Rate & 0.1 & 0.05 & Initial learning rate \\
Epochs & 25 & 30 & Training epochs \\
Word n-grams & 2 & 3 & Maximum length of word n-gram \\
Vector Dimension & 100 & 150 & Embedding dimension \\
Context Window & 5 & 7 & Size of context window \\
Min Word Count & 5 & 3 & Minimum word frequency \\
\bottomrule
\end{tabular}
\caption{Training parameters for FastText language classifiers. The input classifier is optimized for raw text classification, while the output classifier is specifically tuned for transliterated text patterns with slightly different hyperparameters.}
\label{tab:classifier_params}
\end{table*}

\begin{table}[H]
\centering
\begin{tabular}{lrrr}
\toprule
\textbf{Language} & \textbf{Precision} & \textbf{Recall} & \textbf{F1} \\
\midrule
Tibetan & 0.992 & 0.989 & 0.991 \\
Mongolian & 0.987 & 0.985 & 0.986 \\
Uyghur & 0.995 & 0.993 & 0.994 \\
Chinese & 0.998 & 0.997 & 0.998 \\
Other & 0.981 & 0.978 & 0.980 \\
\bottomrule
\end{tabular}
\caption{Performance of the input language classifier on various languages.}
\label{tab:input_classifier_performance}
\end{table}

\subsection{Transliterated Text Classifier}

To accurately identify transliterated text in model outputs and guide proper language restoration, we trained a specialized FastText classifier. The distinguishing feature of this classifier lies in its need to process transliterated text; therefore, we utilized parallel corpora of transliterated text for training, ensuring the model could recognize textual features under different transliteration strategies. The training parameters for the output language classifier are shown in Table~\ref{tab:classifier_params}.

The classification performance on the transliterated text test set is shown in Table~\ref{tab:output_classifier_performance}.

\begin{table}[H]
\centering
\begin{tabular}{lrrr}
\toprule
\textbf{Language} & \textbf{Precision} & \textbf{Recall} & \textbf{F1} \\
\midrule
Tibetan & 0.988 & 0.985 & 0.987 \\
Mongolian & 0.983 & 0.981 & 0.982 \\
Uyghur & 0.991 & 0.989 & 0.990 \\
Chinese & 0.995 & 0.994 & 0.995 \\
Other & 0.992 & 0.990 & 0.991 \\
\bottomrule
\end{tabular}
\caption{Performance of the transliteration text classifier in various languages.}
\label{tab:output_classifier_performance}
\end{table}

\subsection{Pinyin-to-Chinese Converter}

For Chinese pinyin conversion, we performed task-specific fine-tuning based on the Qwen2.5-0.5B model~\cite{yang2024qwen2}. This model handles the conversion from pinyin sequences to Chinese characters, which is a typical sequence-to-sequence conversion task. We used approximately 1 million pinyin-character pairs for training, with data sourced from news texts, Wikipedia, and general domain texts. The parameters used for fine-tuning are shown in Table~\ref{tab:pinyin_converter_params}.

\begin{table}[H]
\centering
\setlength\tabcolsep{12pt}
\begin{tabular}{lrr}
\toprule
\textbf{Parameter} & \textbf{Value} \\
\midrule
Batch Size & 128 \\
Learning Rate & 2e-5 \\
Max Length & 2048 \\
Epochs & 3  \\
Warmup Steps & 1000 \\
Weight Decay & 0.01 \\
\bottomrule
\end{tabular}
\caption{Qwen2.5-0.5B fine-tuning parameter settings.}
\label{tab:pinyin_converter_params}
\end{table}

The performance evaluation of the model on the test set is shown in Table~\ref{tab:pinyin_converter_performance}.

\begin{table*}[t]
\centering
\setlength\tabcolsep{20pt}
\begin{tabular}{lrr}
\toprule
\textbf{Metric} & \textbf{Value} & \textbf{Note} \\
\midrule
Character Accuracy & 0.975 & Single character accuracy \\
Sentence Accuracy & 0.892 & Complete sentence accuracy \\
BLEU Score & 96.8 & Overall translation quality \\
Inference Speed & 125ms/sent & Average processing time \\
\bottomrule
\end{tabular}
\caption{Evaluation of Pinyin Conversion Model Performance. The model shows strong performance in character-level accuracy and complete sentence conversion, with reasonable inference speed suitable for real-time applications.}
\label{tab:pinyin_converter_performance}
\end{table*}


These three auxiliary models collectively form a complete language processing pipeline, capable of accurately identifying input languages, processing transliterated text, and converting pinyin to Chinese characters when needed. In practical applications, these models have demonstrated stable performance and high accuracy.

\section{Compression Analysis}
\label{sec:compression_analysis}

To comprehensively evaluate different transliteration approaches, we compare our three strategies with two baselines: vocabulary expansion (adding 6,000 tokens for each low-resource language) and UROMAN (a widely-used romanization tool). Table \ref{tab:detailed_compression} presents the compression performance across different approaches and languages.

\begin{table}[H]
\small
\centering
\setlength{\tabcolsep}{4pt}
\begin{tabular}{llrr}
\toprule
\multirow{2}{*}{\textbf{Method}} & \multirow{2}{*}{\textbf{Lang}} & \textbf{File} & \textbf{Token} \\
& & \textbf{Compr.} & \textbf{Compr.} \\
\midrule
\multirow{4}{*}{Vocab Expansion} 
& bo & 1.00× & 6.92× \\
& mn & 1.00× & 9.66× \\
& ug & 1.00× & 4.49× \\
& zh & 1.00× & 1.75× \\
\midrule
\multirow{4}{*}{UROMAN} 
& bo & 2.07× & 1.88× \\
& mn & 2.41× & 5.16× \\
& ug & 1.78× & 2.46× \\
& zh & 1.00× & 1.12× \\
\midrule
\multirow{4}{*}{Basic} 
& bo & 1.98× & 1.33× \\
& mn & 1.95× & 2.57× \\
& ug & 1.25× & 1.00× \\
& zh & 0.73× & 0.90× \\
\midrule
\multirow{4}{*}{Tokenizer} 
& bo & 2.61× & 1.80× \\
& mn & 2.07× & 3.85× \\
& ug & 1.27× & 1.39× \\
& zh & 0.73× & 0.90× \\
\midrule
\multirow{4}{*}{Hybrid} 
& bo & 2.61× & 2.23× \\
& mn & 2.07× & 4.98× \\
& ug & 1.27× & 1.92× \\
& zh & 0.73× & 1.41× \\
\bottomrule
\end{tabular}
\caption{Compression performance across different approaches and languages. File Compr. shows the ratio of original file size to transliterated file size. Token Compr. indicates the ratio of original token count to transliterated token count using Llama2 tokenizer. Language codes: bo (Tibetan), mn (Mongolian), ug (Uyghur), zh (Chinese).}
\label{tab:detailed_compression}
\end{table}

The vocabulary expansion approach achieves the highest token compression ratios but maintains original file sizes. UROMAN demonstrates good compression performance in both file size and token count. Our proposed methods show progressive improvements from Basic to Hybrid strategies, with the Hybrid approach achieving competitive token compression while maintaining strong file size reduction. Note that Chinese (zh) shows different patterns due to its unique characteristics in tokenization and encoding.

\section{Training Details}
\label{sec:training_details}
\subsection{Pre-training Data}
The statistics of the raw corpora used for pre-training are shown in Table~\ref{tab:pretraining_data}. The data is primarily sourced from the CUTE parallel corpus \cite{zhuang2025cute}, which provides high-quality aligned multilingual content across Chinese, Uyghur, Tibetan, and English languages. For Mongolian, we followed the data collection and quality assessment methodology described in the CUTE paper to ensure comparable data quality and distribution.

\begin{table}[H]
\centering
\begin{tabular}{lrr}
\toprule
\textbf{Language} & \textbf{Lines} & \textbf{Size (GB)} \\ \midrule
Tibetan (bo) & 934,140 & 11.22 \\
Mongolian (mn) & 933,941 & 11.48 \\
Uyghur (ug) & 934,002 & 7.37 \\
Chinese (zh) & 933,946 & 2.54 \\
English (en) & 933,989 & 3.60 \\ \bottomrule
\end{tabular}
\caption{Pre-training Corpora Statistics. The data is primarily sourced from the CUTE parallel corpus, with additional Mongolian data collected following similar quality standards.}
\label{tab:pretraining_data}
\end{table}

\subsection{Training Parameter Settings}
Table~\ref{tab:training_params} lists the main parameter settings for the pre-training and supervised fine-tuning phases.

\begin{table*}[ht]
\centering
\begin{tabular}{lrr}
\toprule
\textbf{Hyperparameter} & \textbf{Pre-training} & \textbf{Fine-tuning} \\
\midrule
Learning Rate & 1.0e-4 & 2.0e-5 \\
Training Epochs & 1.0 & 3.0 \\
Global Batch Size & 1024 & 256 \\
Max Sequence Length & 4096 & 4096 \\
Warmup Ratio & 0.05 & 0.05 \\
Data Type & BF16 & BF16 \\
LR Scheduler & Cosine & Cosine \\
\bottomrule
\end{tabular}
\caption{Hyperparameter settings for pre-training and supervised fine-tuning phases. During the pre-training phase, except for vocabulary expansion, we observed frequent loss spike phenomena when the learning rate was set to 2.0e-4. After reducing it to 1.0e-4, the training process became more stable.}
\label{tab:training_params}
\end{table*}

\section{Dataset Details}

\subsection{WCM-v2 Dataset}
\label{sec:WCM_v2_details}
WCM-v2 is a multilingual text classification dataset covering 10 domains including arts, geography, and history~\cite{yang2022cino}. The dataset is characterized by its balanced distribution across both categories and languages, containing Chinese training sets and test sets in multiple languages. Table~\ref{tab:wcm_stats} shows the sample distribution of each language across different categories.

\begin{table*}[ht]
\centering
\renewcommand{\arraystretch}{1.2}  
\setlength\tabcolsep{12pt}          
\begin{tabular}{l|rrr|rr}
\toprule
\textbf{Category} & \textbf{mn} & \textbf{bo} & \textbf{ug} & \textbf{zh-train} & \textbf{zh-test} \\
\midrule
Arts        & 135  & 141 & 3   & 2,657  & 335  \\
Geography   & 76   & 339 & 256 & 12,854 & 1,644 \\
History     & 66   & 111 & 0   & 1,771  & 248  \\
Nature      & 7    & 0   & 7   & 1,105  & 110  \\
Natural Science & 779 & 133 & 20  & 2,314  & 287  \\
People      & 1,402 & 111 & 0   & 7,706  & 924  \\
Technology  & 191  & 163 & 8   & 1,184  & 152  \\
Education   & 6    & 1   & 0   & 936    & 118  \\
Economy     & 205  & 0   & 0   & 922    & 109  \\
Health      & 106  & 111 & 6   & 551    & 73   \\
\midrule
Total       & 2,973 & 1,110 & 300 & 32,000 & 3,995 \\
\bottomrule
\end{tabular}
\caption{Sample distribution across categories and languages in the WCM-v2 dataset. The dataset contains training and test sets for Chinese (zh), and test sets for ethnic minority languages (mn: Mongolian, bo: Tibetan, ug: Uyghur). Additional test sets for Korean, Kazakh, and Kyrgyz are also available in the dataset but not used in our experiments.}
\label{tab:wcm_stats}
\end{table*}

\subsection{Machine Reading Comprehension Datasets}
\label{sec:MRC_details}
We utilize three machine reading comprehension (MRC) datasets for evaluation. Table~\ref{tab:mrc_stats} shows the key statistics of these datasets.

\begin{table}[H]
\centering
\begin{tabular}{lrrr}
\toprule
\textbf{Dataset} & \textbf{Train} & \textbf{Dev} & \textbf{Test} \\
\midrule
SQuAD v1.1 & 87,599 & 10,570 & - \\
CMRC 2018 & 10,142 & 3,219 & 1,002 \\
TibetanQA & - & - & 2,007 \\
\bottomrule
\end{tabular}
\caption{Statistics of machine reading comprehension datasets used in our experiments. TibetanQA is used only for testing cross-lingual transfer capability.}
\label{tab:mrc_stats}
\end{table}

\paragraph{SQuAD} The Stanford Question Answering Dataset (SQuAD) v1.1~\cite{rajpurkar2016squad} is a widely used English reading comprehension dataset containing over 100,000 question-answer pairs. The questions and answers were created by crowdworkers based on Wikipedia articles, with answers being continuous spans from the corresponding reading passages.

\paragraph{CMRC} The Chinese Machine Reading Comprehension (CMRC) 2018 dataset~\cite{cui2019span} follows a similar format to SQuAD, featuring span-extraction questions in Chinese. The dataset covers various domains, making it suitable for evaluating Chinese reading comprehension capabilities.

\paragraph{TibetanQA} TibetanQA~\cite{sun2021construction} is a Tibetan machine reading comprehension dataset, with 2,007 publicly released question-answer pairs for evaluation. While the full dataset contains 20,000 question-answer pairs annotated from articles on Tibetan web resources, only a portion is publicly available and used in our experiments for zero-shot cross-lingual evaluation.

\paragraph{Note on Language Coverage} While our study aims to evaluate cross-lingual transfer across multiple low-resource languages, we were unable to identify suitable machine reading comprehension datasets for Mongolian and Uyghur languages at the time of our research. This limitation highlights the scarcity of evaluation resources for these languages in certain NLP tasks.

\section{Translation Details}
\label{sec:translation_details}
\paragraph{Flores-200 Dataset}

The Flores-200 dataset is a multilingual benchmark for evaluating machine translation systems, encompassing 200 languages~\cite{costa2022no}. The sentences in the dataset are derived from English Wikipedia articles and professionally translated into other languages. We utilize both the development and test sets, which contain 997 and 1,012 samples per language pair, respectively. To ensure fair evaluation, we conduct our experiments exclusively on the test set.

\paragraph{Translation Prompts}

We employ English as the unified instruction language for translation tasks. Each prompt contains three carefully selected translation examples (3-shot), with low-resource language text appearing only in the source-target translation pairs. Specifically, the prompt template follows this structure:

\begin{enumerate}[label=(\arabic*)]
    \item An instruction header:
    \begin{center}
    \ttfamily
    Translate the following Chinese text to \{target\_language\}
    \end{center}

    \item Three example translation pairs, each formatted as:
    \begin{center}
    \ttfamily
    Chinese: [source text]\\
    \{target\_language\}: [translation]
    \end{center}

    \item The translation request:
    \begin{center}
    \ttfamily
    Now translate this:\\
    Chinese: [input text]
    \end{center}
\end{enumerate}

This design is motivated by two key considerations: First, utilizing English as the instruction language leverages the model's strong capabilities in English; Second, by minimizing the presence of low-resource languages in the prompt, we can better evaluate the model's genuine translation capabilities rather than simple pattern matching. The three examples are selected to cover diverse sentence structures and vocabulary complexity, helping the model understand the requirements of the translation task.

\end{document}